\documentclass[12pt]{article}
\usepackage{mathrsfs}
\usepackage{natbib,graphicx,setspace,lscape,longtable}
\usepackage{mathrsfs,amsmath,amsthm,amssymb,color}
\usepackage{natbib,epsfig,graphicx,pdfpages}
\usepackage{rotating}
\usepackage{float}
\usepackage{bm}
\usepackage{ulem}
\usepackage{CJK}
\usepackage{natbib}

\usepackage{bbm}
\usepackage{latexsym}
\usepackage{algorithm}

\usepackage{algorithmic}

\bibpunct{(}{)}{;}{a}{,}{,}

\newtheorem{theorem}{Theorem}
\newtheorem{lemma}{Lemma}
\newtheorem{proposition}{Proposition}

\newtheorem{example}{Example}
\def\beq{\begin{equation}}
\def\eeq{\end{equation}}
\def\beqr{\begin{eqnarray}}
\def\eeqr{\end{eqnarray}}
\def\beqrs{\begin{eqnarray*}}
\def\eeqrs{\end{eqnarray*}}
\def\bet{\begin{theorem}}
\def\eet{\end{theorem}}
\def\bel{\begin{lemma}}
\def\eel{\end{lemma}}
\def\bep{\begin{proposition}}
\def\eep{\end{proposition}}
\def\bg{\begin{figure}[tbph]\begin{center}}
\def\eg{\end{center}\end{figure}}

\def\bc{\begin{center}}
\def\ec{\end{center}}

\numberwithin{equation}{section}

\begin{document}
\begin{center}
{\bf\Large Divide and Conquer Local Average Regression}\\
\bigskip

Xiangyu Chang$^1$, Shaobo Lin$^{2*}$ and Yao Wang$^3$
\begin{footnotetext}[1]
{$^*$Correspondence Author: sblin1983@gmail.com
}
\end{footnotetext}

{\it $^1$Center of Data Science and Information Quality, School of Management, Xi'an Jiaotong University, Xi'an, China\\
$^2$Department of Statistics, College of Mathematics and Information Science, Wenzhou University, Wenzhou, China\\
$^3$ Department of Statistics, School of Mathematics and Statistics, Xi'an Jiaotong University, Xi'an, China}\\

This Version: \today

\end{center}

\begin{singlespace}
\begin{abstract}

The divide and conquer strategy, which
breaks a massive data set into a series of manageable data blocks, and then combines the independent results of data blocks to obtain a final decision, has been recognized as a state-of-the-art method to overcome challenges of massive data analysis. In this paper, we merge the divide and conquer strategy with local average regression methods to infer the regressive relationship of input-output pairs from a massive data set. After theoretically analyzing the pros and cons,
we find that although the divide and conquer local average
regression can reach the optimal learning rate, the restriction to
the number of data blocks is a bit strong, which makes it only feasible
for small number of data blocks. We then propose two variants to lessen (or remove) this restriction. Our results show that these variants can
achieve the optimal learning rate with much milder restriction (or
without such restriction). Extensive experimental studies are carried out to verify our theoretical assertions.
\\

\noindent {\bf KEY WORDS: Divide and Conquer Strategy, Local Average Regression, Nadaraya-Watson Estimate,
$k$ Nearest Neighbor Estimate}

\end{abstract}
\end{singlespace}

\newpage

\section{Introduction}

Rapid expansion of capacity in the automatic data generation and
acquisition has made a profound impact on statistics and machine learning, as it
brings data in unprecedented size and complexity. {These data are}
generally called as the ``massive data'' or ``big data''
\citep{Wu2014}. Massive data bring new opportunities of
discovering subtle population patterns and heterogeneities which are
believed to embody rich values and impossible to be found in
relatively small data sets. It, however, simultaneously leads to a
series of challenges such as the storage bottleneck, efficient
computation, etc.~\citep{Zhou2014}.

To attack the aforementioned challenges, a  divide and conquer
strategy was suggested and widely used in statistics and machine
learning communities
\citep{li2013statistical,Mann2009,Zhang2013,Zhang2014,dwork2010differential,battey2015distributed,wang2015statistical}.
This approach firstly distributes a massive data set into $m$ data
blocks, then runs a specified learning algorithm on each data
block independently to get a {\it local estimate} $\hat{f}_{j},
j=1,\dots, m$ and at last transmits $m$ local estimates into one
machine to synthesize a {\it global estimate} $\bar{f}$, which is
expected to model the structure of original massive dataset. A
practical and exclusive synthesizing method is the {\it average
mixture} (AVM)
\citep{li2013statistical,Mann2009,Zhang2013,Zhang2014,dwork2010differential,battey2015distributed},
i.e., $\bar{f} =\frac{1}{m}\sum_{j=1}^m\hat{f}_{j}$.

In practice, the above divide and conquer strategy has many
applicable scenarios. We show the following three situations as
motivated examples. The first one is using limited primary memory to
handle a massive data set. In this situation, the divide and
conquer strategy is employed as a two-stage procedure. In the first
stage, it reads the whole data set sequentially {block by block},
each having a manageable sample size for the limited primary memory,
and derives a {local estimate} based on each  block. In the
second stage, it  averages {local estimates} to build up a global estimate~\citep{li2013statistical}. The second motivated
example is using distributed data management systems to tackle
massive data. In this situation, distributed data management systems
(e.g., Hadoop) are designed by the  divide and conquer strategy.
They can load the whole data set into the systems and tackle
computational tasks separably and automatically. \citet{rhipe2012}
has developed an integrated programing environment of R and Hadoop
(called RHIPE) for expedient and efficient statistical computing. The third
motivated example is the massive data privacy. In this situation, it
divides a massive data set into several small pieces and combining
the estimates derived from these pieces for keeping the data
privacy~\citep{dwork2010differential}.

For nonparametric regression, the aforementioned
 divide and conquer strategy has been shown to be efficient and
feasible for global modeling methods such as the kernel ridge
regression \citep{Zhang2014} and conditional maximum entropy model
\citep{Mann2009}. Compared with the global modeling methods, local
average regression (LAR)~\citep{gyorfi2006distribution,fan2000prospects,tsybakov2008introduction}, such as the Nadaraya-Watson kernel (NWK) and $k$ nearest neighbor (KNN)
 estimates, benefits in computation and therefore, is also widely used in image processing~\citep{takeda2007kernel}, power
prediction~\citep{kramer2010power}, recommendation
system~\citep{biau2010statistical} and financial
engineering~\citep{Kato2012}. LAR is by definition a learning scheme
that averages outputs whose corresponding inputs  satisfy certain
localization assumptions. To tackle massive data  regression
problems, we combine the divide and conquer approach with LAR to
produce a new learning scheme, average mixture local average
regression (AVM-LAR), just as \cite{Zhang2014} did for kernel ridge
regression.

Our first purpose is to analyze the performance of AVM-LAR. After
providing the optimal learning rate of LAR,  we show that AVM-LAR
can also achieve this rate, provided the number of data blocks, $m$,
is relatively small. We also prove that the restriction
concerning $m$ cannot be essentially improved. In a word, we provide
both the optimal learning rate and the essential restriction
concerning $m$ to guarantee the optimal rate of AVM-LAR. It should
be highlighted that this essential restriction is a bit strong and
makes AVM-LAR feasible only for small $m$. Therefore, compared with LAR, AVM-LAR does not bring the essential improvement, since we must pay much attention to
determine an appropriate $m$.

Our second purpose is to pursue other divide and conquer strategies to equip LAR efficiently. In particular, we provide two concrete variants of AVM-LAR in this
paper. The first variant is motivated by the distinction between KNN
and NWK. In our experiments, we note that the range of $m$ to
guarantee the optimal learning rate of AVM-KNN is much larger than
AVM-NWK. Recalling that the localization parameter of KNN depends
on data, while it doesn't hold for NWK, we propose a variant of
AVM-LAR such that localization parameters of each data block depend on data. We establish the optimal learning
rate of this variant with  mild restriction to $m$ and verify its
feasibility by numerical simulations.
 The second variant is based
on the definitions of AVM and LAR. It follows from the definition of
LAR that the predicted value of a new input depends  on samples near
the input. If there are not such samples in a specified data block,
then this data block doesn't affect the prediction of LAR. However,
AVM averages {local estimates} directly, neglecting the concrete
value of a specified {local estimate}, which may lead to an
inaccurate prediction. Based on this observation, we propose another
variant of AVM-LAR by distinguishing whether a specified data block
affects the prediction. We provide the optimal learning rate of this
variant without any restriction to $m$ and also present the
experimental verifications.

To complete the above missions, the rest of paper is organized as
follows. In Section \ref{section2}, we  present optimal learning
rates of LAR and  AVM-LAR and analyze the pros and cons of
AVM-LAR. In Section \ref{section3}, we propose two new modified
AVM-LARs to improve the performance of AVM-LAR. A set of simulation
studies to support the correctness of our assertions are given in
Section \ref{section4}. In Section \ref{section5}, we
detailedly justify all the theorems. In Section \ref{section6}, we present the conclusion and some useful remarks. 

\section{Divide and Conquer Local Average Regression}\label{section2}

In this section, after introducing some basic concepts of LAR, we
present a baseline of our analysis, where an optimal minimax
learning rate of LAR is derived. Then, we deduce the  learning rate
  of AVM-LAR and analyze its pros and cons.

\subsection{Local Average Regression}

Let $D=\{(X_i,Y_i)\}_{i=1}^N$ be the data set where $X_i\in\mathcal
X\subseteq\mathbb{R}^d$ is a covariant and $Y_i\in [-M,M]$ is the
real-valued response. We always assume $\mathcal X$ is a compact
set. Suppose that samples are drawn independently and
identically according to an unknown joint distribution $\rho$ over
$\mathcal X\times [-M,M]$. Then the main aim of nonparametric
regression is to construct a function $\hat{f}:\mathcal X\rightarrow
[-M,M]$ that can describe future responses based on new inputs. The
quality of the estimate $\hat{f}$ is measured in terms of the {\it
mean-squared prediction error} $\mathbf E\{\hat{f}(X)-Y\}^2$, which is
minimized by the so-called {\it regression function}
$f_{\rho}(x)=\mathbf E\left\{Y|X=x\right\}$.

LAR, as one of the most widely used
nonparametric regression approaches, constructs an estimate formed
as
\begin{equation}\label{local_estimation}
f_{D,h}(x)=\sum_{i=1}^NW_{h,X_i}(x)Y_i,
\end{equation}
where the localization weight $W_{h,X_i}$ satisfies $W_{h,X_i}(x)>0$ and
$\sum_{i=1}^NW_{h,X_i}(x)=1$. Here, $h>0$ is the so-called
localization parameter. Generally speaking, $W_{h,X_i}(x)$ is ``small'' if $X_i$
is ``far'' from $x$. Two widely used  examples of LAR are
Nadaraya-Watson kernel (NWK) and $k$ nearest neighbor (KNN)
estimates.
\begin{example}(NWK estimate)
Let $K: \mathcal X\rightarrow\mathbb R_+$ be a kernel function~\citep{gyorfi2006distribution}, and $h>0$ be its bandwidth. The NWK estimate is defined by
\begin{equation}\label{Watson kernel estimate}
            \hat{f}_h(x)=\frac{\sum_{i=1}^NK\left(\frac{x-X_i}{h}\right)Y_i}{\sum_{i=1}^NK\left(\frac{x-X_i}{h}\right)},
\end{equation}
and therefore,
$$
            W_{h,X_i}(x)=\frac{
            K\left(\frac{x-X_i}{h}\right)}{\sum_{i=1}^NK\left(\frac{x-X_i}{h}\right)}.
$$
It is worth noting that we use the convention $\frac{0}{0}=0$ in the following.
Two popular kernel functions are the naive kernel,
$K(x)={I}_{\{\|x\|\leq 1\}}$ and Gaussian kernel
$K(x)=\exp\left(-\|x\|^2\right)$, where $ I_{\{\|x\|\leq 1\}}$ is an indicator function with the feasible domain $\|x\|\leq 1$ and $\|\cdot\|$ denotes the Euclidean norm.
\end{example}
\begin{example}(KNN estimate)
For $x\in \mathcal{X}$, let $\{(X_{(i)}(x),Y_{(i)}(x))\}_{i=1}^N$ be
a permutation of $\{(X_i,Y_i)\}_{i=1}^N$ such that
$$
            \|x-X_{(1)}(x)\|\leq\cdots\leq \|x-X_{(N)}(x)\|.
$$
Then the KNN estimate is defined by
\begin{equation}\label{KNNE estimate}
         \hat{f}_k(x)=\frac1k\sum_{i=1}^k Y_{(i)}(x).
\end{equation}
According to (\ref{local_estimation}), we have
$$
          W_{h,X_i}(x)=\left\{\begin{array}{cc} 1/k, &
          \mbox{if}\  X_i\in\{X_{(1)},\dots,X_{(k)}\},\\
            0, & \mbox{otherwise.}
            \end{array}\right.
$$
Here we denote the weight of KNN  as $W_{h,X_i}$ instead of
$W_{k,X_i}$ in order to unify the notation and $h$ in KNN  depends
  on the distribution of points and $k$.
\end{example}

\subsection{Optimal Learning Rate of LAR}

The weakly universal consistency and optimal learning rates of
some specified LARs have been justified by~\citet{stone1977consistent,stone1980optimal,stone1982optimal} and summarized in the book \citet{gyorfi2006distribution}. In particular, \citet[Theorem
4.1]{gyorfi2006distribution} presented a sufficient condition to
guarantee the weakly universal consistency of LAR. \citet[Theorem
5.2, Theorem 6.2]{gyorfi2006distribution} deduced  optimal learning
rates of NWK  and KNN.
The aim of this subsection is to present some sufficient conditions to guarantee optimal learning rates of general LAR.

For $r,c_0>0$, let $\mathcal
F^{c_0,r}=\{f|f:\mathcal{X}\rightarrow\mathcal{Y},|f(x)-f(x')|\leq
c_0\|x-x'\|^r,\forall x, x'\in\mathcal{X}\}$. We suppose in this
paper that $f_{\rho} \in \mathcal F^{c_0,r}$. This is a commonly
accepted prior assumption of regression function which is employed
in
~\citep{tsybakov2008introduction,gyorfi2006distribution,Zhang2014}.
The following  Theorem \ref{THEOREM:SUFFICIENT CONDITION FOR LOCAL
ESTIMATE} is our first main result.

\begin{theorem}\label{THEOREM:SUFFICIENT CONDITION FOR LOCAL ESTIMATE}
Let $f_{D,h}$ be defined by (\ref{local_estimation}).
Assume that:

(A) there exists a positive number $c_1$ such that
$$
       \mathbf E\left\{\sum_{i=1}^NW^2_{h,X_i}(X)\right\}\leq
       \frac{c_1}{Nh^d};
$$

(B) there exists a positive number  $c_2$  such that
$$
      \mathbf
      E\left\{ \sum_{i=1}^NW_{h,X_i}(X)I_{\{\|X-X_i\|>h\}}\right\}
      \leq
       \frac{c_2}{\sqrt{Nh^d}}.
$$
If $h\sim N^{-1/(2r+d)}$, then there exist  constants $C_0$ and
$C_1$ depending only on $d$, $r$, $c_0$, $c_1$ and $c_2$ such that
\begin{equation}\label{theorem1}
           C_0N^{-2r/(2r+d)}\leq \sup_{f_\rho\in\mathcal
           F^{c_0,r}}\mathbf E\{\|f_{D,h}-f_\rho\|_\rho^2\}
           \leq
           C_1N^{-2r/(2r+d)}.
\end{equation}
\end{theorem}

Theorem \ref{THEOREM:SUFFICIENT CONDITION FOR LOCAL ESTIMATE}
presents sufficient conditions of the localization weights to ensure the
optimal learning rate of LAR. There are totally four constraints
of the weights $W_{h,X_i}(\cdot)$. The first one is the averaging
constraint $\sum_{i=1}^NW_{h,X_i}(x)=1, \ \mbox{for all}\
X_i,x\in\mathcal
      X.$ It essentially reflects the word ``average'' in LAR.
The second one is the non-negative constraint. We regard it as a
mild constraint  as it holds for all the widely used LAR such as
NWK and KNN. The third constraint is  condition (A), which devotes
to controlling the scope of the weights. It aims at avoiding the
extreme case that there is a very large weight near 1 and others are
almost 0. The last constraint is condition (B), which implies the
localization property of LAR.


We should highlight that Theorem \ref{THEOREM:SUFFICIENT CONDITION
FOR LOCAL ESTIMATE} is significantly  important for our analysis. On
the one hand, it is obvious that the AVM-LAR estimate (see Section \ref{AVM-LARsection}) is also a new LAR estimate. Thus, Theorem
\ref{THEOREM:SUFFICIENT CONDITION FOR LOCAL ESTIMATE} provides a
theoretical tool to derive  optimal learning rates of AVM-LAR. On
the other hand, Theorem \ref{THEOREM:SUFFICIENT CONDITION FOR LOCAL
ESTIMATE} also provides a sanity-check that an efficient AVM-LAR
estimate should possess the similar learning rate as
(\ref{theorem1}).


\subsection{AVM-LAR}\label{AVM-LARsection}

The AVM-LAR estimate, which is a marriage of the classical AVM
strategy \citep{Mann2009,Zhang2013,Zhang2014} and LAR, can be formulated in the  following Algorithm \ref{AVM-LAR}.

\begin{algorithm}\caption{AVM-LAR}\label{AVM-LAR}
\begin{algorithmic}[!h]
\STATE {{\bf Initialization}: Let $D=\{(X_i,Y_i)\}_{i=1}^N$ be $N$
samples, $m$ be the number of data blocks, $h$ be the bandwidth
parameter.}

\STATE{ {\bf Output}: The global estimate $\overline{f}_h$.}

\STATE{{\bf Division}:  Randomly divide $D$ into $m$ data blocks
$D_1,D_2,\dots,D_m$ such that $D=\mathop{\bigcup}\limits_{j=1}^{m}
D_j, D_i\cap D_j=\varnothing, i\neq j$ and
$|D_1|=\dots=|D_m|=n=N/m$.}

 \STATE{{\bf Local
processing}:
           For  $j=1,2,\dots,m$, implement LAR for the
data block $D_j$ to get the $j$th {\it local estimate}
$$
        f_{j,h}(x)=\sum_{(X_i,Y_i)\in D_j}W_{X_i,h}(x)Y_i.
$$
}

\STATE{{\bf Synthesization}: Transmit   $m$ {\it local estimates}
$f_{j,h}$ to a machine, getting a {\it global estimate} defined by
\begin{equation}\label{global_RERM}
            \overline{f}_h=\frac{1}{m}\sum_{j=1}^mf_{j,h}.
\end{equation}}
\end{algorithmic}
\end{algorithm}

In Theorem \ref{THEOREM DISTRBUTED LAE POSITIVE}, we show that this
simple generalization of LAR achieves the optimal learning rate
with a rigorous condition concerning $m$. We also show that this condition is essential.


\begin{theorem}\label{THEOREM DISTRBUTED LAE POSITIVE}
Let $\overline{f}_h$ be defined by (\ref{global_RERM}) and $h_{D_j}$
 be the mesh norm of the data block $D_j$ defined by $
                 h_{D_j}:=\max\limits_{X\in\mathcal X}\min\limits_{X_i\in
                 D_j}\|X-X_i\|.
$
 Suppose that

(C) for all $D_1,\dots,D_m$, there exists a positive number $c_3$
such that
$$
       \mathbf E\left\{\sum_{(X_i,Y_i)\in D_j}W^2_{h,X_i}(X)\right\}\leq
       \frac{c_3}{nh^d};
$$

(D) for all $D_1,\dots,D_m$, there holds almost surely
$$
          W_{X_i,h}I_{\{\|x-X_i\|>h\}}=0.
$$
If $h\sim N^{-1/(2r+d)}$, and the
event \{$ h_{D_j}\leq h$ for all $D_j$\} holds, then there exists a constant $C_2$ depending only on $d,r,M,c_0,c_3$ and $c_4$ such that 
\begin{equation}\label{theorem1.11111}
           C_0N^{-2r/(2r+d)}\leq \sup_{f_\rho\in\mathcal
           F^{c_0,r}}\mathbf E\{\|\overline{f}_{h}-f_\rho\|_\rho^2\}
           \leq
           C_2N^{-2r/(2r+d)}.
\end{equation}
Otherwise, if the event \{$ h_{D_j}\leq h$ for all $D_j$\} dose not
hold, then for arbitrary $h\geq\frac12(n+2)^{-1/d}$,  there exists a
distribution  $\rho$ such that
\begin{equation}\label{theorem1.22222222222}
            \sup_{f_\rho\in\mathcal
           F^{c_0,r}}\mathbf E\{\|\overline{f}_{h}-f_\rho\|_\rho^2\}
           \geq \frac{M^2\{(2h)^{-d}-2\}}{3n}.
\end{equation}

\end{theorem}

The assertions in Theorem \ref{THEOREM DISTRBUTED LAE
POSITIVE} can be divided into two parts. The first one is the
positive assertion, which means that if some  conditions of the
weights and  an extra constraint of the data blocks are imposed, then
the AVM-LAR estimate (\ref{global_RERM}) possesses the same learning
rate as that in (\ref{theorem1}) by taking the same localization
parameter $h$ (ignoring constants). This means that the proposed
 divide and conquer operation in
 Algorithm \ref{AVM-LAR} doesn't affect the learning rate under this circumstance.  In fact,
we can relax the restriction (D) for the bound
(\ref{theorem1.11111}) to the following condition (D$^*$).

 (D$^*$)  For all $D_1,\dots,D_m$, there exists a positive number
$c_4$ such that
$$
      \mathbf
      E\left\{\sum_{(X_i,Y_i)\in D_j}W_{h,X_i}(X)I_{\{\|X-X_i\|>h\}}\right\}
      \leq
       \frac{c_4}{\sqrt{Nh^d}}.
$$
To guarantee the optimal minimax learning rate of AVM-LAR, condition
(C) is the same as   condition (A) by noting that there are only $n$
samples in each $D_j$. Moreover,  condition (D$^*$) is a bit
stronger than condition (B) as there are totally $n$ samples in
$D_j$ but the localization bound of it is $c_4/(\sqrt{Nh^d})$.
However, we should point out that such a restriction is also  mild,
since in almost all   widely used LAR, the localization bound either
is  0 (see NWK with naive kernel, and KNN) or decreases
exponentially (such as NWK with Gaussian kernel). All the above
methods satisfy conditions (C) and (D$^*$).

The negative assertion, however, shows that if the event \{there is
a $D_j$ such that $h_{D_j}>h$\} holds, then for any
$h\geq\frac12(n+2)^{-1/d}$, the learning rate of AVM-LAR isn't
faster than $\frac{1}{nh^d}$. It follows from Theorem
\ref{THEOREM:SUFFICIENT CONDITION FOR LOCAL ESTIMATE} that the best
localization parameter to guarantee the optimal  learning rate
satisfies {$h\sim N^{-1/(2r+d)}$}. The condition
$h\geq\frac12(n+2)^{-1/d}$ implies that if the best parameter is
selected, then $m$ should satisfy $m\leq \mathcal O(N^{2r/(2r+d)})$.
Under this condition, from
  (\ref{theorem1.22222222222}), we have
$$
         \sup_{f_\rho\in\mathcal
           F^{c_0,r}}\mathbf E\{\|\overline{f}_{h}-f_\rho\|_\rho^2\}
           \geq  \frac{C}{nh^d}.
$$
This means, if we select $h\sim N^{-1/(2r+d)}$ and $m\leq \mathcal
O(N^{2r/(2r+d)})$, then the learning rate of AVM-LAR is essentially
slower than that in (\ref{theorem1}). If we select a smaller $h$,
then the above inequality also yields the similar conclusion. If we
select a larger $h$, however, the
 approximation error (see the proof of Theorem \ref{THEOREM:SUFFICIENT CONDITION FOR LOCAL ESTIMATE})
 is $\mathcal O(h^{2r})$ which
can be larger than the learning rate in (\ref{theorem1}). In a word,
if the event \{$ h_{D_j}\leq h$ for all $D_j$\} does not hold, then
the AVM operation essentially degrades the optimal learning rate of
LAR.

At last, we should discuss the probability of the event \{$
h_{D_j}\leq h$ for all $D_j$\}. As $\mathbf P \{
      h_{D_j}\leq h  \ \mbox{for all}\  D_j \} =1-m\mathbf
      P\{h_{D_1}>h\},$
 and it can   be found in \citep[P.93-94]{gyorfi2006distribution} that
$\mathbf P\{h_{D_1}>h\} \leq \frac{c}{nh^d}$, we have $\mathbf P \{
      h_{D_j}\leq h  \ \mbox{for all}\  D_j \} \geq
      1-\frac{m}{nh^d}.$
When $h\sim (mn)^{-1/(2r+d)}$, we have
$$
      \mathbf P \{
      h_{D_j}\leq h  \ \mbox{for all}\  D_j \} \geq
      1-c' \frac{m^{(2r+2d)/(2r+d)}}{n^{2r/(2r+d)}}.
$$
The above quantity is small when $m$ is large, which means
 that the event \{$ h_{D_j}\leq h$ for all $D_j$\} has a significant
chance to be broken down.
 By
using the method in \citep[Problem 2.4]{gyorfi2006distribution}, we
can  show that  the above estimate for the confidence is essential
in the sense that for the uniform distribution, the equality holds
for some constant $c'$.

\section{Modified AVM-LAR}\label{section3}

As shown in Theorem \ref{THEOREM DISTRBUTED LAE POSITIVE}, if
$h_{D_j}\leq h$ does not hold for some $D_j$, then AVM-LAR cannot
reach the optimal learning rate. In this section, we propose two
variants of AVM-LAR such that they can achieve the optimal learning
rate under mild conditions.

\subsection{AVM-LAR with data-dependent parameters}

The event \{$ h_{D_j}\leq h$ for all $D_j$\} essentially implies
that for arbitrary $x$, there is at least one sample in the ball
$B_{h}(x):=\{x'\in \mathbb{R}^d: \|x-x'\|\leq h$\}. This condition holds for KNN  as the parameter $h$ in KNN changes
with respect to  samples. However, for NWK and other local average
methods (e.g., partition estimation~\citep{gyorfi2006distribution}), such a condition usually fails.  Motivated by KNN, it is
natural to select a sample-dependent localization $h$ to ensure the
event \{$ h_{D_j}\leq h$ for all $D_j$\}. Therefore, we propose a
variant of AVM-LAR with data-dependent parameters in Algorithm
\ref{AVM-LAR-alg2}.

\begin{algorithm}\caption{AVM-LAR with data-dependent parameters}\label{AVM-LAR-alg2}
\begin{algorithmic}[!h]
\STATE {{\bf Initialization}: Let $D=\{(X_i,Y_i)\}_{i=1}^N$ be $N$
samples, $m$ be the number of data blocks.}

\STATE{ {\bf Output}: The global estimate $\tilde{f}_{\tilde{h}}$.}

\STATE{{\bf Division}: Randomly divide $D$ into $m$ data blocks
$D_1,D_2,\dots,D_m$ such that $D=\mathop{\bigcup}\limits_{j=1}^{m}
D_j, D_i\cap D_j=\varnothing, i\neq j$ and $|D_1|=\dots=|D_m|=n=N/m$.
Compute the mesh norms $h_{D_1},\dots, h_{D_m}$, and select
$\tilde{h}\geq h_{D_j}$, $j=1,2,\dots,m$.
  }
\STATE{{\bf Local processing}:
           For any $j=1,2,\dots,m$, implement   LAR with bandwidth parameter $\tilde{h}$ for the
data block $D_j$ to get the $j$th {\it local estimate}
$$
        f_{j,\tilde{h}}(x)=\sum_{(X_i,Y_i)\in D_j}W_{X_i,\tilde{h}}(x)Y_i.
$$
}

\STATE{{\bf Synthesization}: Transmit   $m$ {\it local estimates}
$f_{j,\tilde{h}}$ to a machine, getting a {\it global estimate}
defined by
\begin{equation}\label{global_AVM_LAR1}
            \tilde{f}_{\tilde{h}}=\frac{1}{m}\sum_{j=1}^mf_{j,\tilde{h}}.
\end{equation}}
\end{algorithmic}
\end{algorithm}

Comparing with AVM-LAR in Algorithm \ref{AVM-LAR}, the only
difference of Algorithm \ref{AVM-LAR-alg2} is  the division step,
where we select the bandwidth parameter to be  greater than all
$h_{D_j}, j =1,\dots,m$. The following  Theorem \ref{THEOREM:
AVM-LAE} states the theoretical merit of AVM-LAR with
data-dependent parameters.

\begin{theorem}\label{THEOREM: AVM-LAE}
Let $r<d/2$, $\tilde{f}_{\tilde{h}}$ be defined by
(\ref{global_AVM_LAR1}). Assume (C) and (D$^*$) hold. Suppose
$$
\tilde{h}=\max\{m^{-1/(2r+d)}\max_j\{h_{D_j}^{d/(2r+d)}\},\max_j\{h_{D_j}\}\},$$
and
$m\leq\left(\frac{
c_0^2(2r+d)+8d(c_3+2c_4^2)M^2}{4r(c_0^2+2)}\right)^{d/(2r)}N^{2r/(2r+d)},
$ then there exists a constant $C_3$ depending only on $c_0,c_3,c_4,r,d$ and $M$ such that
\begin{equation}\label{theorem3}
           C_0N^{-2r/(2r+d)}\leq \sup_{f_\rho\in\mathcal
           F^{c_0,r}}\mathbf E\{\|\tilde{f}_{\tilde{h}}-f_\rho\|_\rho^2\}
           \leq
           C_3N^{-2r/(2r+d)}.
\end{equation}
\end{theorem}

Theorem \ref{THEOREM: AVM-LAE} shows that if the localization parameter
is selected elaborately, then AVM-LAR can achieve the optimal
learning rate under mild conditions concerning $m$. It should be
noted that there is an additional restriction to the smoothness
degree, $r<d/2$. We highlight that this condition cannot be removed.
In fact, without this condition, (\ref{theorem3}) may not hold for
some marginal distribution $\rho_X$. For example, let $d=1$, it can
be deduced from \citep[Problem 6.1]{gyorfi2006distribution} that
there exists a $\rho_{X}$ such that (\ref{theorem3}) doesn't hold.
However, if we don't aim at deriving a distribution free result, we
can fix this condition by using the technique in \citep[Problem
6.7]{gyorfi2006distribution}. Actually, for $d\leq 2r$,  assume the
marginal distribution $\rho_X$ satisfies that there exist
$\varepsilon_0>0$, a nonnegative function $g$ such that for all
$x\in\mathcal X$, and $0<\varepsilon\leq \varepsilon_0$ satisfying
$\rho_X(B_\varepsilon(x))>g(x)\varepsilon^d,$ and $\int_{\mathcal
         X}\frac{1}{g^{2/d}(x)}d\rho_X<\infty,$
then (\ref{theorem3}) holds for arbitrary $r$ and $d$. It is obvious
that the   uniform distribution satisfies the above conditions.

Instead of imposing a restriction to $h_{D_j}$, Theorem
\ref{THEOREM: AVM-LAE} states that after using the data-dependent
parameter $\tilde{h}$, AVM-LAR doesn't degrade the learning rate
for a large range of $m$. We should illustrate that the derived
bound of $m$ cannot be improved further. Indeed, it can be found in
our proof that  the bias of AVM-LAR can be bounded by $C\mathbf
E\{\tilde{h}^{2r}\}$. Under the conditions of Theorem \ref{THEOREM:
AVM-LAE}, if $m\sim N^{(2r+\varepsilon)/(2r+d)}$, then for arbitrary
$D_j$, there holds
$\mathbf E\{ {h_{D_j}}\}\leq Cn^{-1/d}
      = C(N/m)^{-1/d}\leq CN^{(d-\varepsilon)/(2r+d)}.$
Thus, it is easy to check that $\mathbf E\{\tilde{h}^{2r}\}\leq
CN^{(-2r+\varepsilon)/(2r+d)}$, which implies a learning rate slower
than $N^{-2r/(2r+d)}$.

\subsection{Qualified AVM-LAR}

Algorithm \ref{AVM-LAR-alg2} provided an intuitive way to improve
the performance of AVM-LAR. However,  Algorithm \ref{AVM-LAR-alg2}
  increases the computational complexity of AVM-LAR, because we
have to compute the mesh norm $h_{D_j},j=1,\dots, m$. A natural
question is whether we can avoid this procedure while maintaining
the learning performance. The following Algorithm \ref{AVM-LAR-alg3}
provides a possible way to tackle this question.

\begin{algorithm}\caption{Qualified AVM-LAR}\label{AVM-LAR-alg3}
\begin{algorithmic}
\STATE {{\bf Initialization}: Let $D=\{(X_i,Y_i)\}_{i=1}^N$ be $N$
samples, $m$ be the number of data blocks, $h$ be the bandwidth parameter.}

\STATE{ {\bf Output}: The global estimate $\hat{f}_h$.}

\STATE {{\bf Division}: Randomly divide $D$ into $m$ data blocks,
i.e. $D=\cup_{j=1}^mD_j$ with $D_j\cap D_k=\varnothing$ for $k\neq
j$ and $|D_1|=\dots=|D_m|=n$.}

\STATE {{\bf Qualification}: For a test input $x$, if there exists
an $X_{0}^j\in D_j$ such that $|x-X_0^j|\leq h,$ then we qualify
$D_j$ as an active data block for the {\it local estimate}. Rewrite
all the active data blocks as $T_1,\dots,T_{m_0}$.}

\STATE {{\bf Local processing }:  For arbitrary data block $T_j$,
$j=1,\dots,m_0$, define
$$
           f_{j,h}(x)=\sum_{(X_i,Y_i)\in T_j}W_{X_i,h}(x)Y_i.
$$
}

\STATE{{\bf Synthesization}: Transmit   $m_0$ {\it local estimates}
$f_{j,{h}}$ to a machine, getting a {\it global estimate} defined by
\begin{equation}\label{New AVM}
            \hat{f}_h=\frac1{m_0}\sum_{j=1}^{m_0}f_{j,h}.
\end{equation}
}

\end{algorithmic}
\end{algorithm}

Comparing with Algorithms \ref{AVM-LAR} and \ref{AVM-LAR-alg2}, the
only difference of Algorithm \ref{AVM-LAR-alg3} is the qualification
step which essentially doesn't need extra   computation. In fact,
the qualification and local processing steps can be implemented
simultaneously.  It should be further mentioned that the
qualification step actually eliminates   the data blocks which have
a chance to break down the event \{$h_{D_j}\leq h$ for all $D_j$\}.
Although, this strategy may loss a  part of information, we show
that the qualified AVM-LAR can achieve the optimal learning rate
without any restriction to $m$.

\begin{theorem}\label{THEOREM: DLAE}
Let $\hat{f}_{h}$ be defined by (\ref{New AVM}).  Assume
(C) holds and

(E) for all $D_1,\dots,D_m$, there exists a positive number
$c_5$ such that
$$
      \mathbf
      E\left\{ \sum_{i=1}^n|W_{h,X_i}(X)|I_{\{\|X-X_i\|>h\}}\right\}
      \leq
       \frac{c_{5}}{m\sqrt{nh^d}}.
$$
If $h\sim N^{-1/(2r+d)}$, then there exists a constant $C_4$
depending only on $c_0,c_1,c_3,c_5,r,d$ and $M$ such that
\begin{equation}\label{theorem dlae}
           C_0N^{-2r/(2r+d)}\leq \sup_{f_\rho\in\mathcal
           F^{c_0,r}}\mathbf E\{\|\hat{f}_{h}-f_\rho\|_\rho^2\}
           \leq
           C_4N^{-2r/(2r+d)}.
\end{equation}
\end{theorem}

In Theorem \ref{THEOREM: AVM-LAE}, we declare that AVM-LAR with
data-dependent parameter doesn't slow down  the learning rate of
LAR. However, the bound of $m$ in Theorem \ref{THEOREM: AVM-LAE}
depends on the smoothness of the regression function, which is
usually unknown in the real world applications. This makes $m$ be a
potential parameter in AVM-LAR with data-dependent parameter, as we
do not know which $m$ definitely works. However, Theorem
\ref{THEOREM: DLAE} states that we can avoid this problem by
introducing a qualification step. The theoretical price of such an
improvement is only to use  condition (E) to take place condition
(D$^*$). As shown above, all the widely used LARs such as the
partition estimate, NWK with naive kernel, NWK  with Gaussian kernel
and KNN satisfy condition (E) (with a logarithmic term for NWK
with Gaussian kernel).

\section{Experiments}\label{section4}
In this section, we report experimental studies on synthetic data
sets to demonstrate the
performances of  AVM-LAR and its variants. We employ three criteria
for the comparison purpose. The first criterion is the {\it global
error} (GE) which is the mean square error of testing set when $N$
samples are used as a training set. We use  GE as a baseline that
does not change with respect to $m$. The second criterion is the
{\it local error} (LE) which is the mean square error of testing set
when we use only one data block ($n$ samples) as a training set. The
third criterion is the {\it average error} (AE) which is the mean
square error of AVM-LAR (including Algorithms \ref{AVM-LAR},
\ref{AVM-LAR-alg2} and \ref{AVM-LAR-alg3}).
\subsection{Simulation 1}
We use a fixed total number of samples
$N=10,000$, but assume that the number of data blocks $m$ (the data block size
$n=N/m$) and dimensionality $d$ are varied. The
simulation results are based on the average values of 20 trails.

We generate data from the following regression models $
            y = g_j(x)+\varepsilon, \ j=1,2,
$ where $\varepsilon$ is the Gaussian noise $\mathcal{N}(0,0.1)$,
\begin{equation}\label{simulation1}
g_1(x)=\left\{
\begin{array}{cc}
(1-2x)_+^3(1+6x), & 0<x\leq 0.5 \\
0 & x>0.5%
\end{array}%
\right.,
\end{equation}
and
\begin{equation}\label{simulation2}
g_2(x)=\left\{
\begin{array}{cc}
(1-\|x\|)_+^5(1+5\|x\|) + \frac{1}{5}\|x\|^2, & 0<\|x\|\leq 1, x\in\mathbb{R}^5 \\
 \frac{1}{5}\|x\|^2 & \|x\|>1%
\end{array}%
\right..
\end{equation}
\citet{Wendland2005} revealed that $g_1$ and $g_2$ are the so-called
Wendland functions with the property $g_1,g_2\in
\mathcal{F}^{c_0,1}$ and $g_1,g_2\notin \mathcal{F}^{c_0,2}$ for
some absolute constant $c_0$. The simulated $N$ samples are drawn
i.i.d. according to the uniform distribution on the (hyper-)cube
$[0,1]^d$. We also generate 1000 test samples $(X_i',Y_i')$ with
$X_i'$ drawn i.i.d. according to the uniform distribution and
$Y_i'=g_j(X_i'), j=1,2$.

\begin{figure*}[ht]
\begin{center}
  \includegraphics[width=450pt]{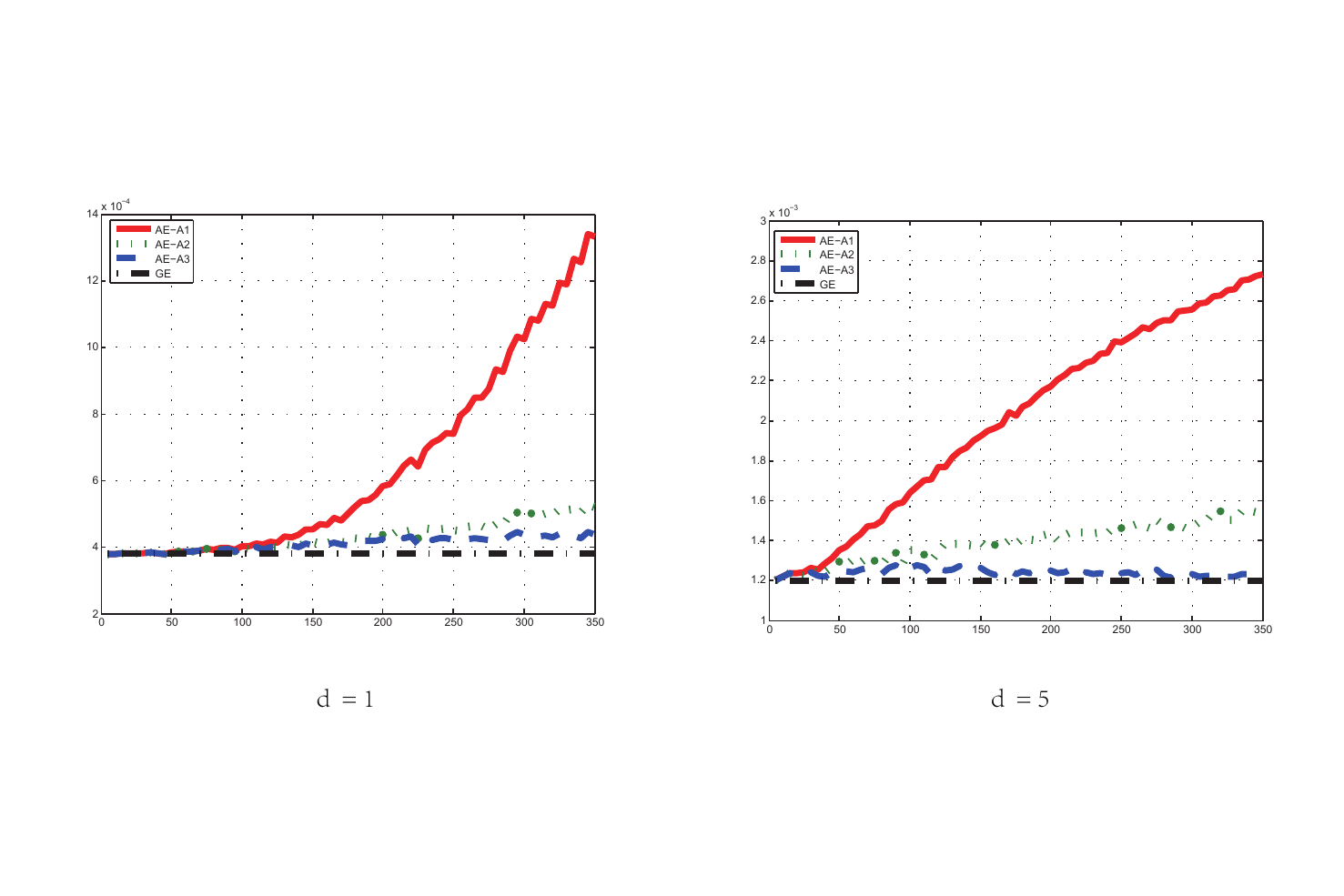}
  \vspace{-25mm}
\caption{\emph{AE-A1, AE-A2, AE-A3 and GE of the simulation. The curves of AE-A2 and AE-A3 are always below AE-A1's to illustrate the improved capability of modified AVM-LARs.}}\label{simulations2_NWK}
  \end{center}
\end{figure*}

On the basis of above setting,  we illustrate two
simulation results. The first one is to compare the learning
performance between Algorithm \ref{AVM-LAR} and LAR. Both NWK  and
KNN  are  considered. The second one is to show how  Algorithms
\ref{AVM-LAR-alg2} and \ref{AVM-LAR-alg3}   overcome Algorithm
\ref{AVM-LAR}'s weakness. Because Algorithms \ref{AVM-LAR},
\ref{AVM-LAR-alg2} and \ref{AVM-LAR-alg3} are the same for KNN, KNN is not considered in this part. The detailed implementation of
NWK and KNN is specified as follows.
\begin{itemize}

\item NWK: In Algorithm 1
and Algorithm \ref{AVM-LAR-alg3}, for each $m\in\{5,10,\dots,350\}$,  the bandwidth
 parameter satisfies  $h\sim N^{-\frac{1}{2r+d}}$ according to
 Theorem \ref{THEOREM DISTRBUTED LAE POSITIVE} and Theorem \ref{THEOREM: DLAE}.
In Algorithm \ref{AVM-LAR-alg2}, we set $\tilde{h}\sim
 \max\{m^{-1/(2r+d)}\max_j\{h_{D_j}^{d/(2r+d)}\},\max_j\{h_{D_j}\}\}$
according to Theorem \ref{THEOREM: AVM-LAE}.

\item KNN: According to Theorem \ref{THEOREM DISTRBUTED LAE POSITIVE}, the parameter $k$ is set to $k\sim\frac{N^{\frac{2r}{2r+d}}}{m}$. However,
as $k\geq 1$,  the range of $m$ should satisfy
$m\in\{1,2,\dots,N^{\frac{2r}{2r+d}}\}$.
\end{itemize}
To present   proper constants for the localization parameters (e.g.,
$h=cN^{-\frac{1}{2r+d}}$), we use the 5-fold cross-validation method in simulations. Based on these strategies, we obtain the simulation results in the following Figures \ref{simulations1_NWK}, \ref{simulations1_KNN} and \ref{simulations2_NWK}.

As shown in  Figure \ref{simulations1_NWK}, AEs are smaller than
LEs, which means  AVM-NWK outperforms NWK  with only one data block.
Furthermore, AEs of NWK are comparable with GEs when $m$ is not
too big and  there exists an upper bound  of the number of data
blocks, $m'$, lager than which  the curve of AE increases
dramatically. Moreover, $m'$ decreases when $d$ increases.

\begin{figure*}[ht]
\begin{center}
  \includegraphics[width=400pt]{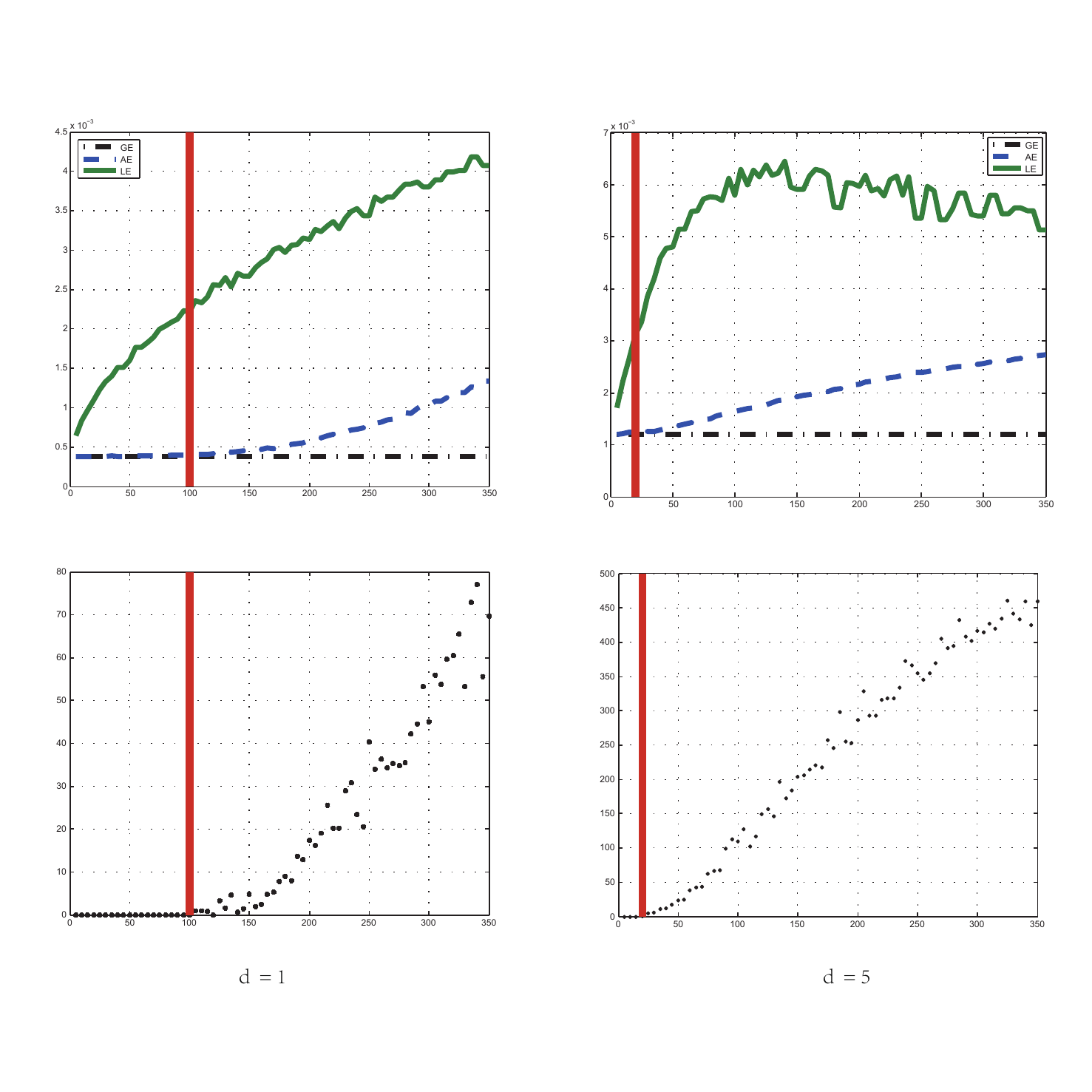}
  \vspace{-15mm}
\caption{\emph{The first row shows AEs, LEs and GEs of NWK for
different $m$. The second row shows the number of inactive machines
which satisfy $h_{D_j}>h$. The vertical axis of the second row of
Figure \ref{simulations1_NWK} is the number of inactive data blocks
which break  down the condition $h_{D_j}\leq h$.
}}\label{simulations1_NWK}
  \end{center}
\end{figure*}

Let us explain these above phenomena. If only one data block is
utilized, then it follows from Theorem \ref{THEOREM:SUFFICIENT
CONDITION FOR LOCAL ESTIMATE} that $\min\limits_{j=1,\dots,m}\mathbf
E\{\|{f}_{j,h}-f_\rho\|_\rho^2\}=\mathcal{O}(n^{-\frac{2r}{2r+d}})$,
which is far larger than $\mathcal{O}(N^{-\frac{2r}{2r+d}})$ for
AVM-NWK due to Theorem \ref{THEOREM DISTRBUTED LAE POSITIVE}. Thus,
AEs are  smaller than LEs.  Moreover, Theorem \ref{THEOREM
DISTRBUTED LAE POSITIVE}  implies that AEs are comparable with GE as
long as the  event \{$ h_{D_j}\leq h$ for all $D_j, j=1,\dots, m$\}
holds. To verify this assertion, we record the number of data blocks
breaking down the condition $h_{D_j}\leq h$ for different $m$ in the
second row of Figure \ref{simulations1_NWK}. It can be observed that the
dramatically increasing time of the number of inactive data blocks
and AEs are almost same. This result is extremely consistent with
the negative part of Theorem \ref{THEOREM DISTRBUTED LAE POSITIVE}.
Because   large $d$ and  $m$ lead to a higher probability to break
down the event \{$ h_{D_j}\leq h$ for all $D_j, j=1,\dots, m$\}.
Thus   $m'$ decreases when $d$ increases.

\begin{figure*}[ht]
\begin{center}
 \includegraphics[width=430pt]{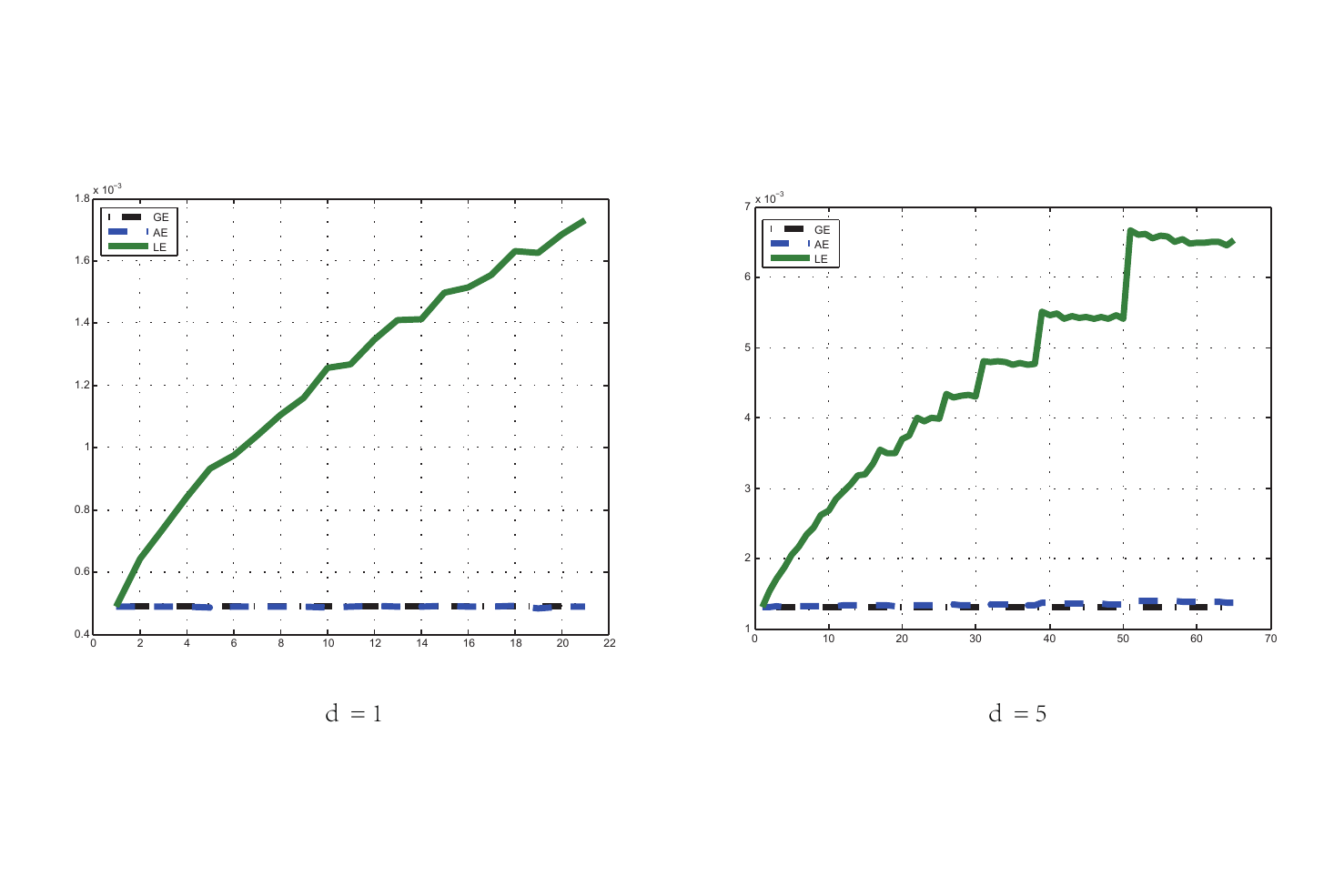}
  \vspace{-15mm}
\caption{\emph{ AEs, LEs and GEs of KNN for different
$m$.}}\label{simulations1_KNN}
  \end{center}
\end{figure*}

Compared with AVM-NWK, AVM-KNN shows significant different results
in Figure \ref{simulations1_KNN}. In fact, there isn't a similar
$m'$ to guarantee comparable AEs and GE for AVM-KNN. The reason is
   KNN selects a data-dependent bandwidth $h$ which makes the
event \{$ h_{D_j}\leq h$ for all $D_j, j=1,\dots, m$\} always holds.
This result is extremely consistent with the positive part of
Theorem \ref{THEOREM DISTRBUTED LAE POSITIVE}. However, we find that
AVM-KNN has a design deficiency. To be detailed, the range of $m$
must be in $\{1,2,\dots,N^{\frac{2r}{2r+d}}\}$ due to $k\geq 1$.

In Figure \ref{simulations2_NWK}, AEs of Algorithms \ref{AVM-LAR},
\ref{AVM-LAR-alg2} and \ref{AVM-LAR-alg3} which are denoted by
AE-A1, AE-A2 and AE-A3. We can find that AE-A1, AE-A2 and AE-A3 have
similar  values which are comparable with GE when $m\leq m'$. The
reason is that the small number of data blocks can guarantee the
event \{$h_{D_j}\leq h$ for all $D_j, j=1,\dots,m$\}. Under this
circumstance, Theorems \ref{THEOREM DISTRBUTED LAE POSITIVE},
\ref{THEOREM: AVM-LAE} and \ref{THEOREM: DLAE} yield that all these
  estimates can reach  optimal learning rates. As $m$ increasing,
the event \{$h_{D_j}>h$ for some $j$\} inevitably happens, then
Algorithm \ref{AVM-LAR} fails according to the negative part of
Theorem \ref{THEOREM DISTRBUTED LAE POSITIVE}. At the same time,
AE-A1 begins to increase dramatically. As Algorithms
\ref{AVM-LAR-alg2} and \ref{AVM-LAR-alg3}  are designed  to avoid
the weakness of Algorithm \ref{AVM-LAR}, the curves of AE-A2 and
AE-A3 are always below that of AE-A1 when $m>m'$. Another
interesting fact is that AE-A3 is smaller than AE-A2,
although both of them all can achieve the same learning rate in  theory.

\subsection{Simulation 2}
We make use of the same simulation study which is conducted by
\citet{Zhang2014} for comparing the learning performance of AVM-NWK
(including Algorithms \ref{AVM-LAR}, \ref{AVM-LAR-alg2} and
\ref{AVM-LAR-alg3}) and the divide and conquer kernel ridge regression
(DKRR for short).

We generate data from the regression model $y=g_3(x)+\epsilon$,
where $g_3(x)=\min(x,1-x)$, the noise variable $\epsilon$ is
normally distributed with mean 0 and variance $\sigma^2=1/5$, and
$X_i,i=1,\dots,N$ are simulated from a uniform distribution in
$[0,1]$ independently. In \citet{Zhang2014}'s simulation, DKRR used
the kernel function $K(x,x')=1+\min\{x,x'\}$, and regularization
parameter $\lambda = N^{-2/3}$ due to $g_3\in \mathcal{F}^{c_0,1}$
for some absolute constant $c_0$. We also use $N=10,000$ training
samples, and 1,000 test samples. The parameter setting of Algorithms
\ref{AVM-LAR}, \ref{AVM-LAR-alg2} and \ref{AVM-LAR-alg3} is the same
as that in Simulation 1.

\vspace{5mm}
\begin{figure*}[ht]
\begin{center}
  \includegraphics[width=300pt]{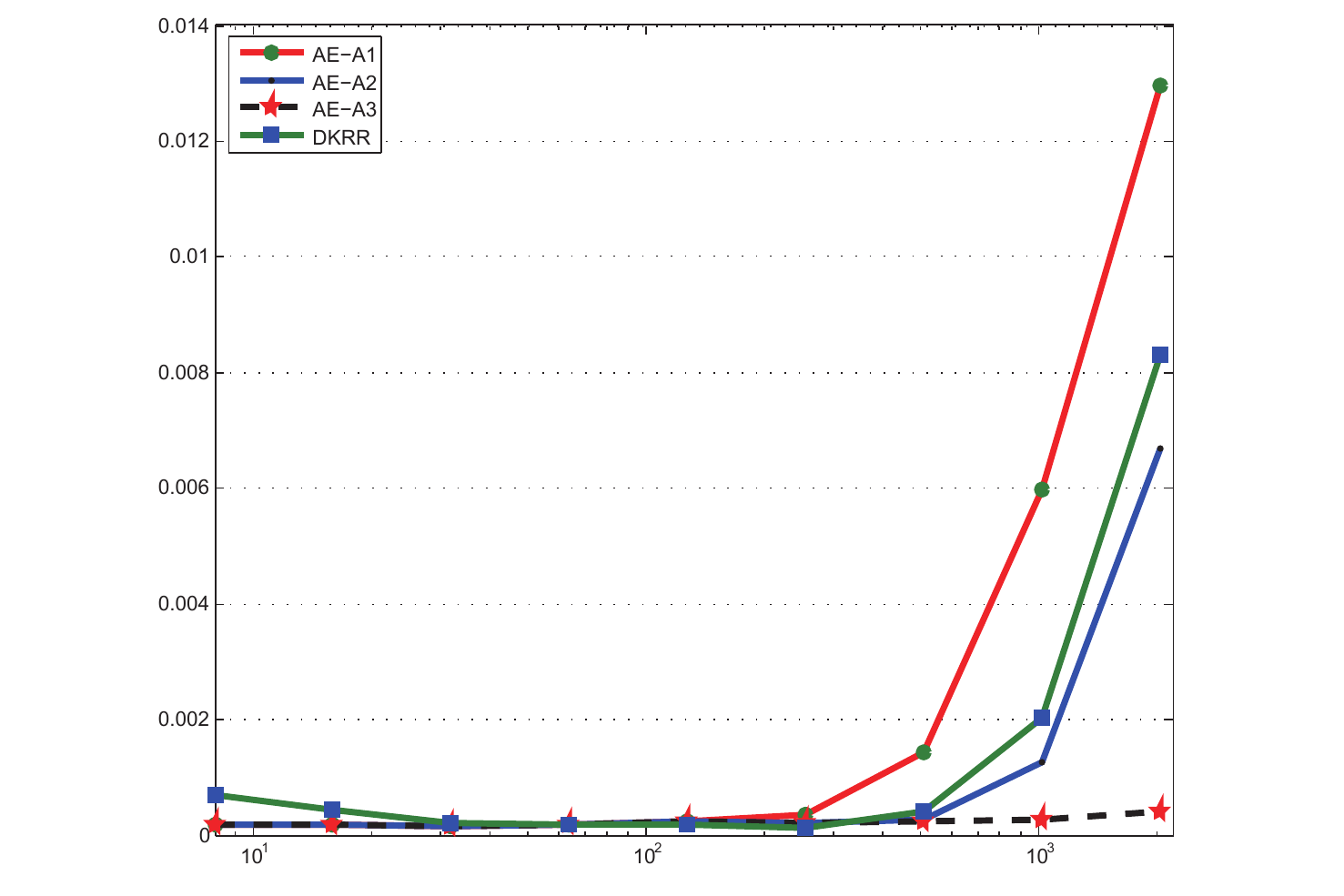}

\caption{\emph{AEs of Algorithm \ref{AVM-LAR},  \ref{AVM-LAR-alg2},
\ref{AVM-LAR-alg3} and DKRR.
}}\label{simulations2_com}
  \end{center}
\end{figure*}

In Figure \ref{simulations2_com}, we plot AEs of Algorithms
\ref{AVM-LAR}, \ref{AVM-LAR-alg2}, \ref{AVM-LAR-alg3} and DKRR for
$m\in\{2^3,2^4,\dots, 2^{11}\}$. Figure \ref{simulations2_com} shows
AEs of Algorithm \ref{AVM-LAR}, \ref{AVM-LAR-alg2},
\ref{AVM-LAR-alg3} and DKRR are comparable when $m<256$. As long as
$m>256$, AEs of Algorithm \ref{AVM-LAR}, \ref{AVM-LAR-alg2} and DKRR
increase dramatically. However, AEs of Algorithm \ref{AVM-LAR-alg3}
are  stable. The reason is that, to keep the optimal learning rates,
DKRR needs $m = \mathcal{O}(N^{1/3})$~\citep{Zhang2014}, and
Algorithm \ref{AVM-LAR-alg2} needs $m=\mathcal{O}(N^{2/3})$, while
Algorithm \ref{AVM-LAR-alg3} holds for all $m$.

\subsection{3D Road Network Data}

Building accurate 3D road networks is one of centrally important
problems for {\it Advanced Driver Assistant Systems} (ADAS).
Benefited  from an accurate 3D road network, eco-routing, as an
application of ADAS, can yield fuel cost savings 8-12\% when
compared with standard routing~\citep{tavares2009optimisation}. For this reason, obtaining an accurate 3D road networks plays an important role for ADAS~\citep{kaul2013building}.

North Jutland (NJ), the northern part of Justland, Denmark, covers a
region of 185km$\times$130km. NJ contains a spatial road network
with a total length of $1.17\times 10^7m$, whose 3D ployline
representation is containing 414,363 points. Elevation values where
extracted from a publicly available massive Laser Scan Point Clod
for Denmark are added in the data set. Thus, the data set includes 4
attribute: {\it OSMID} which is the OpenStreetMap ID for each road
segment or edge in the graph; {\it longitude and latitude} with
Google format; {\it elevation} in meters.  In practice, the acquired
data set may include missing values. In this subsection, we try to
use AVM-LAR based on Algorithms \ref{AVM-LAR}, \ref{AVM-LAR-alg2}
and \ref{AVM-LAR-alg3} for rebuilding the missing elevation
information on the points of 3D road networks via aerial laser scan
data.

\begin{figure*}[ht]
\begin{center}
 \includegraphics[width=330pt]{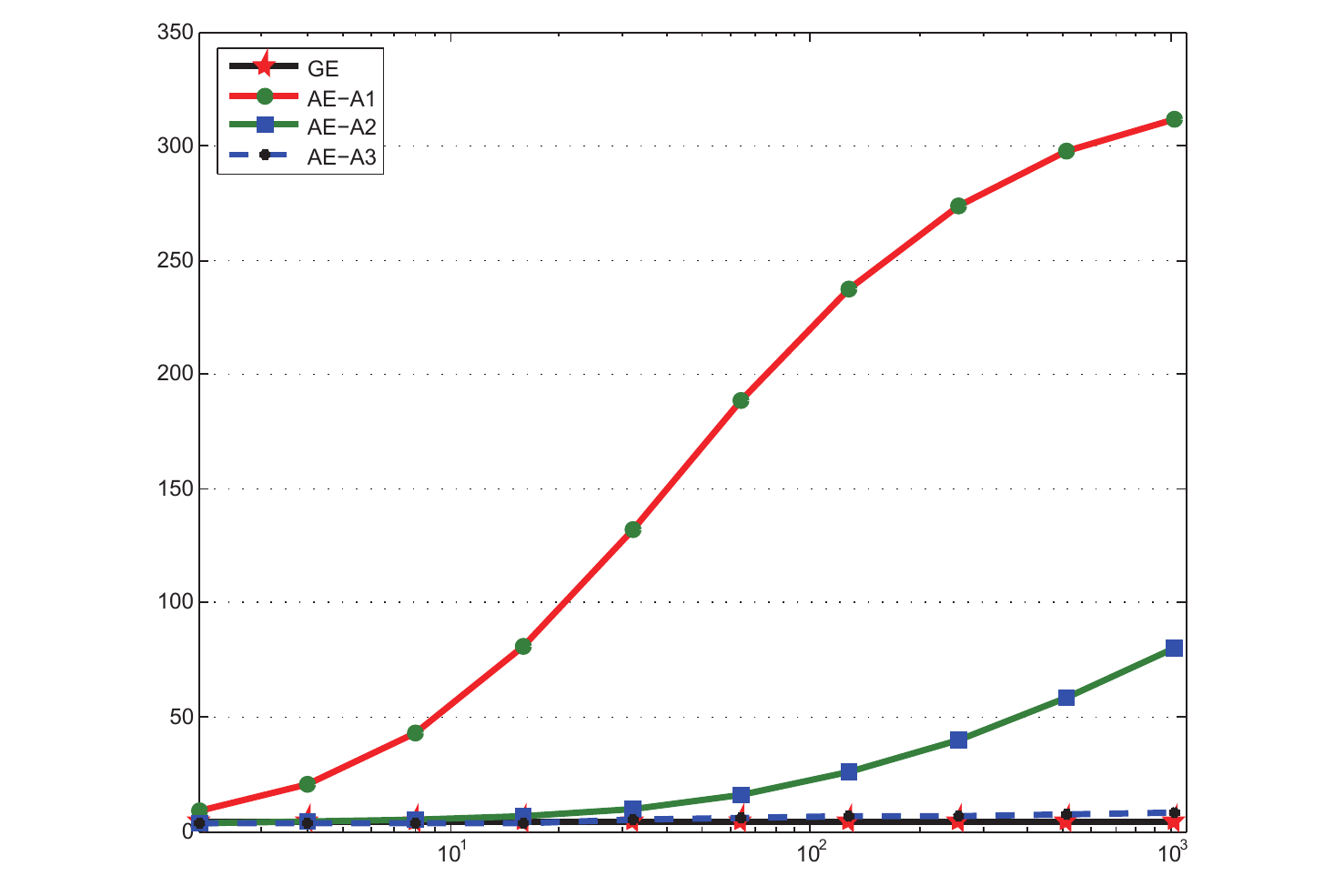}
\caption{\emph{AE-A1, AE-A2, AE-A3 and GE of the 3D road network
data.}}\label{figure_road}
  \end{center}
\end{figure*}

To this end, we randomly select 1000 samples as a test set  (record
time seed for the reproducible research). Using the other samples,
we run AVM-NWK based on Algorithm \ref{AVM-LAR}, \ref{AVM-LAR-alg2}
and \ref{AVM-LAR-alg3} to predict the missed elevation information
in the test set. Here, the bandwidth $h=0.13N^{-1/4}$ and
$N=413,363$.  AE-A1, AE-A2, AE-A3 and GE for different
$m\in\{2,2^2,\dots,2^{10}\}$ are recorded in Figure
\ref{figure_road}.

\vspace{5mm}

We can find in Figure \ref{figure_road} that AEs of Algorithm
\ref{AVM-LAR} are  larger than GE, which implies the weakness of
AVM-NWK based on the negative part of Theorem \ref{THEOREM
DISTRBUTED LAE POSITIVE}. AE-A3 has  almost same values with the GE
for all $m$, however, AE-A2 possesses similar  property only when
$m\leq 32.$ Then, for the 3D road network data set,  Algorithm
\ref{AVM-LAR-alg3} is applicable to fix the missed elevation
information for the data set.

\section{Proofs}\label{section5}

\subsection{Proof of Theorem 1}
Let $f_{\rho,h}(x)= \sum_{i=1}^NW_{h,X_i}(x)f_\rho(X_i).$
Then, it is obvious that $f_{\rho,h}(x)=\mathbf E^*\{f_{D,h}(x)\},$
where $\mathbf  E^*\{\cdot\}=\mathbf E\{\cdot|X_1,X_2,\dots,X_n\}$.
Therefore, we can deduce
$$
        \mathbf E^*\{(f_{D,h}(x)-f_\rho(x))\}=\mathbf E^*
        \{(f_{D,h}(x)-f_{\rho,h}(x))^2\}+(f_{\rho,h}(x)-f_\rho(x))^2.
$$
That is,
$$
     \mathbf E\{\|f_{D,h}-f_\rho\|_\rho^2\}
     =
     \int_{\mathcal X}\mathbf E\{\mathbf E^*
        \{(f_{D,h}(X)-f_{\rho,h}(X))^2\}\}d\rho_X+\int_{\mathcal X}
        \mathbf E\{(f_{\rho,h}(X)-f_\rho(X))^2\}d\rho_X.
$$
The first and second terms are referred to the {\it sample error} and {\it
 approximation error}, respectively. To bound the sample error, noting
$\mathbf E^*\{Y_i\}=f_\rho(X_i)$,   we have
\begin{eqnarray*}
           &&\mathbf E^*
        \{(f_{D,h}(x)-f_{\rho,h}(x))^2\}
          =
         \mathbf E^*
        \left\{\left(\sum_{i=1}^NW_{h,X_i}(x)(Y_i-f_\rho(X_i))\right)^2
        \right\}\\
        &\leq&
         \mathbf E^*\left\{\sum_{i=1}^N\left(W_{h,X_i}(x)(Y_i-f_\rho(X_i))\right)^2
        \right\}
        \leq
         4M^2\sum_{i=1}^NW^2_{h,X_i}(x).
\end{eqnarray*}
Therefore we can use (A) to bound the sample error as
$$
        \mathbf E\{(f_{D,h}(X)-f_{\rho,h}(X))^2\}\leq4M^2\mathbf E
        \left\{\sum_{i=1}^NW^2_{h,X_i}(X)\right\}\leq
        \frac{4c_1M^2}{Nh^d}.
$$
Now, we turn to bound the approximation error. Let $B_h(x)$   be the
$l^2$ ball with center $x$ and radius $h$, we have
\begin{eqnarray*}
        &&\mathbf E\{(f_{\rho,h}(X)-f_\rho(X))^2\}
        =
        \mathbf
        E\left\{\left(\sum_{i=1}^NW_{h,X_i}(X)f_\rho(X_i)-f_\rho(X)\right)^2\right\}\\
        &=&
        \mathbf
        E\left\{\left(\sum_{i=1}^NW_{h,X_i}(X)(f_\rho(X_i)-f_\rho(X))\right)^2\right\}\\
         &=&
        \mathbf
        E\left\{\left(\sum_{i=1}^NW_{h,X_i}(X)(f_\rho(X_i)-f_\rho(X))
        \right)^2I_{\{B_h(X)\cap D=\varnothing\}}
        \right\}\\
        &+&
        \mathbf E\left\{\left(\sum_{i=1}^NW_{h,X_i}(X)(f_\rho(X_i)
        -f_\rho(X))\right)^2I_{\{B_h(X)\cap D\neq\varnothing\}}
        \right\}.
\end{eqnarray*}
It follows from \citep[P.66]{gyorfi2006distribution} and
$\sum_{i=1}^NW_{h,X_i}(X)=1$ that
$$
         \mathbf E\left\{\left(\sum_{i=1}^NW_{h,X_i}(X)(f_\rho(X_i)
        -f_\rho(X))\right)^2I_{\{B_h(X)\cap D=\varnothing\}}
        \right\}
        \leq \frac{16M^2}{Nh^d}.
$$
Furthermore,
\begin{eqnarray*}
        &&\mathbf
        E\left\{\left(\sum_{i=1}^NW_{h,X_i}(X)(f_\rho(X_i)-f_\rho(X))
        \right)^2I_{\{B_h(X)\cap D\neq\varnothing\}}
        \right\}\\
        &\leq&
        \mathbf
        E\left\{\left(\sum_{\|X_i-X\|\leq h}W_{h,X_i}(X)|f_\rho(X_i)-f_\rho(X)|
        \right)^2I_{\{B_h(X)\cap D\neq\varnothing\}}
        \right\}\\
        &+&
        \mathbf
        E\left\{\left(\sum_{\|X_i-X\|> h}W_{h,X_i}(X)|f_\rho(X_i)-f_\rho(X)|
        \right)^2
        \right\}\\
        &\leq&
        c_0^2h^{2r}+\frac{4c^2_2M^2}{Nh^d},
\end{eqnarray*}
where the last inequality is deduced by $f_\rho\in\mathcal
F^{c_0,r}$, condition (B) and Jensen's inequality.
 Under this circumstance, we get
$$
     \mathbf E\{\|f_{D,h}-f_\rho\|_\rho^2\}
     \leq
     c_0^2h^{2r}+\frac{4(c_1+c_2^2+4)M^2}{Nh^d}.
$$
If we set $h=\left(\frac{4(c_1+c_2^2+4)M^2}{c_0^2N}\right)^{-1/(2r+d)},$
then
$$
           \mathbf E\{\|f_{D,h}-f_\rho\|_\rho^2\}\leq
           c_0^{2d/(2r+d)}(4(c_1+c_2^2+4)M^2)^{2r/(2r+d)}N^{-2r/(2r+d)}.
$$
This together with \citep[Theorem 3.2]{gyorfi2006distribution} finishes the proof
of Theorem \ref{THEOREM:SUFFICIENT CONDITION FOR LOCAL ESTIMATE}. $\square$

\subsection{Proof of Theorem 2}
Since $\mathbf E\left\{\|\overline{f}_h-f_\rho\|_\rho^2\right\}
     =
     \mathbf E\left\{\|\overline{f}_h-\mathbf
     E\{\overline{f}_h\}+\mathbf
     E\{\overline{f}_h\}-f_\rho\|^2_\rho\right\}$
and
$\mathbf E\{ \overline{f}_h \}=\mathbf
          E\{ f_{j,h}\}, \quad j=1,\dots,m,$
we get
\begin{eqnarray}\label{distribution error decomposition}
           \mathbf E\left\{\|\overline{f}_h-f_\rho\|^2_\rho\right\}
            &=&
            \frac1{m^2}\mathbf E\left\{\sum_{j=1}^m\left(\|f_{j,h }-\mathbf
            E\{f_{j,h }\}\|_\rho^2+\|\mathbf
            E\{f_{j,h }\}-f_\rho\|^2_\rho\right)\right. \nonumber \\
            &+&
            \left.2\sum_{j=1}^m\sum_{k\neq j}\langle f_{j,h}-\mathbf
            E\{f_{j,h}\},f_{k,h}-\mathbf
            E\{f_{k,h}\}\rangle_\rho\right\} \nonumber\\
            &=&
           \frac1m\mathbf E\{\| f_{1,h} -\mathbf
            E\{f_{1,h}\}\|^2_\rho\}+ \|\mathbf
            E\{f_{1,h}\}-f_\rho\|^2_\rho\\
            &\leq&
            \frac2m\mathbf E\{\| f_{1,h} -f_\rho\}\|^2_\rho\}+ 2\|\mathbf
            E\{f_{1,h}\}-f_\rho\|^2_\rho. \nonumber
\end{eqnarray}
As $h\geq h_{D_j}$ for all $1\leq j\leq m$, we obtain that
$B_h(x)\cap D_j\neq\varnothing$ for all $x\in\mathcal X$ and $1\leq
j\leq m$. Then, using the same method as that in the proof of
Theorem \ref{THEOREM:SUFFICIENT CONDITION FOR LOCAL ESTIMATE}, (C)
and (D$^*$) yield  that
$$
          \mathbf E\{\| f_{1,h} -f_\rho\}\|^2_\rho\}
          \leq c_0^2h^{2r}+\frac{4(c_3+c_4^2)M^2}{nh^d}.
$$
Due to the Jensen's inequality, we have
$$
       \mathbf E\left\{\|\overline{f}_h-f_\rho\|_\rho^2\right\}
       \leq
       \frac{2c_0^2h^{2r}}{m}+\frac{8(c_3+c_4^2)M^2}{mnh^d}+
       2\mathbf E\left\{\|\mathbf
            E^*\{f_{1,h}\}-f_\rho\|^2_\rho\right\}.
$$
Noting $B_h(X)\cap D_j\neq\varnothing$ almost surely, the same
method as that in the proof of Theorem \ref{THEOREM:SUFFICIENT
CONDITION FOR LOCAL ESTIMATE} together with (D) yields that
$$
\mathbf E\left\{\|\mathbf
            E^*\{f_{1,h}\}-f_\rho\|^2_\rho\right\}\leq
            c_0^2h^{2r}+\frac{8c_4^2M^2}{mnh^d}.
$$
Thus,
$$
            \mathbf E\left\{\|\overline{f}_h-f_\rho\|_\rho^2\right\}
            \leq
           \frac{16(c_3+c_4^2)M^2}{mnh^d}
            +3c_0^2h^{2r}.
$$
This finishes the proof of (\ref{theorem1.11111}) by taking  $
h=\left(\frac{16(c_3+c_4^2)M^2}{3c_0^2nm}\right)^{-1/(2r+d)}$ into
account.

 Now, we turn to prove (\ref{theorem1.22222222222}). According to (\ref{distribution error
decomposition}), we have
$$
           \mathbf E\left\{\|\overline{f}_h-f_\rho\|^2_\rho\right\}
           \geq
            \|\mathbf
            E\{f_{1,h}\}-f_\rho\|^2_\rho
            =\int_{\mathcal X}\left(\mathbf
            E\left\{\sum_{i=1}^nW_{X_i,h}(X)f_\rho(X_i)-f_\rho(X)\right\}\right)^2
            d\rho_X.
$$
Without loss of generality, we assume $h<h_{D_1}$. It then follows
from the definition of the mesh norm that   there exits
$X\in\mathcal X$ which is not in $B_h(X_i)$, $X_i\in D_1$. Define the
separation radius of a set of points
$S=\{\zeta_i\}_{i=1}^n\subset\mathcal{X}$
  via
$$
                 q_{_{S}}:=\frac12\min_{j\neq k}\|\zeta_j-\zeta_k\|.
$$
   The mesh ratio $
               \tau_{_{S}}:=\frac{h_{S}}{q_{_{S}}}\geq1
$ provides a measure of how uniformly points in $S$ are distributed
on $\mathcal X$. If $\tau_{_{S}}\leq 2$, we then call $S$ as the
quasi-uniform point set. Let $\Xi_l=\{\xi_1,\dots,\xi_l\}$ be
$l=\lfloor(2h)^{-d}\rfloor$ quasi-uniform points \citep{Wendland2005}
in $\mathcal X$. That is
$\tau_{_{\Xi_l}}=\frac{h_{_{\Xi_l}}}{q_{_{\Xi_l}}}\leq 2.$ Since
$h_{_{\Xi_l}}\geq l^{-1/d}$, we have $q_{_{\Xi_l}}\geq
\frac1{2l^{1/d}}\geq h.$ Then,
\begin{eqnarray}\nonumber
 \mathbf
           E\left\{\|\overline{f}_h-f_\rho\|^2_\rho\right\}
           &=&
           \|\mathbf E\{f_{1,h}\}-f_\rho\|_\rho^2\\\nonumber
           &\geq&
            \sum_{k=1}^l\int_{B_{q_{_{\Xi_l}}}(\xi_k)}\left(\mathbf
            E\left\{\sum_{i=1}^nW_{X_i,h}(X)f_\rho(X_i)-f_\rho(X)\right\}\right)^2
            d\rho_X.
\end{eqnarray}

{If $f_\rho(x)=M$, then
\begin{eqnarray*}
           \mathbf E\left\{\|\overline{f}_h-f_\rho\|^2_\rho\right\}
          & \geq&
            M^2\sum_{k=1}^l\int_{B_{q_{_{\Xi_l}}}(\xi_k)}\left(\mathbf
            E\left\{I_{\{D_1\cap B_{q_{_{\Xi_l}}}(\xi_k)=\varnothing\}}\right\}\right)^2
            d\rho_X\\
            &\geq&
            M^2\sum_{k=1}^l\rho_X(B_{q_{_{\Xi_l}}}(\xi_k))\mathbf P\{D_1\cap
            B_{q_{_{\Xi_l}}}(\xi_k)=\varnothing\}\\
            &=&
            M^2\sum_{k=1}^l\rho_X(B_{q_{_{\Xi_l}}}(\xi_k))(1-\rho_X(B_{q_{_{\Xi_l}}}(\xi_k)))^n.
\end{eqnarray*}
Since $h\geq\frac12(n+2)^{-1/d}$, we can let $\rho_X$ be the
marginal distribution satisfying
$$
         \rho_X(B_{q_{_{\Xi_l}}}(\xi_k))=1/n, \ k=1,2,\dots, l-1.
$$
Then
$$
           \mathbf E\left\{\|\overline{f}_h-f_\rho\|^2_\rho\right\}
           \geq
           M^2\sum_{k=1}^{l-1}\frac1n(1-1/n)^n
           \geq\frac{M^2((2h)^{-d}-2)}{3n}.
$$
  This finishes the proof of Theorem \ref{THEOREM DISTRBUTED LAE
  POSITIVE}. $\square$
}
\subsection{Proof of Theorem 3}
Without loss of generality, we assume $h_{D_1}=\max_j\{h_{D_j}\}$.
It follows from (\ref{distribution error decomposition}) that
$$
           \mathbf E\left\{\|\tilde{f}_{\tilde{h}}-f_\rho\|^2_\rho\right\}
            \leq
            \frac2m\mathbf E\{\| f_{1,\tilde{h}} -f_\rho\}\|^2_\rho\}+ 2\|\mathbf
            E\{f_{1,\tilde{h}}\}-f_\rho\|^2_\rho.
$$
We first bound $\mathbf E\{\| f_{1,\tilde{h}} -f_\rho\}\|^2_\rho\}$.
As $\tilde{h}\geq h_{D_1}$, the same method as that in the proof of
Theorem \ref{THEOREM DISTRBUTED LAE POSITIVE} yields that
$$
        \mathbf E\{\| f_{1,\tilde{h}} -f_\rho\}\|^2_\rho\}
        \leq
        c_0^2\mathbf E\{\tilde{h}^{2r}\}+\mathbf E\left\{
        \frac{4M^2(c_3+c_4^2)}{n\tilde{h}^d}\right\}.
$$
To bound $\|\mathbf
            E\{f_{1,\tilde{h}}\}-f_\rho\|^2_\rho$, we use the same
            method as that in the proof of Theorem \ref{THEOREM DISTRBUTED LAE POSITIVE}
again. As $\tilde{h}\geq m^{-1/(2r+d)}h_{D_1}^{d/(2r+d)}$, it is
easy to deduce that
\begin{eqnarray*}
          \|\mathbf
            E\{f_{1,\tilde{h}}\}-f_\rho\|^2_\rho
            &\leq&
            \mathbf E\left\{\|\mathbf
            E^*\{f_{1,\tilde{h}}\}-f_\rho\|^2_\rho\right\}
            \leq
            c_0^2\mathbf E\{\tilde{h}^{2r}\}+\mathbf
            E\left\{\frac{8c_4^2M^2}{mn\tilde{h}^d}\right\}\\
            &\leq&
            c_0^2m^{-2r/(2r+d)}\mathbf E\{h_{D_1}^{2rd/(2r+d)}\}
            +
            c_0^2\mathbf E\{h_{D_1}^{2r}\}\\
            &+&
            8c_4^2M^2(mn)^{-1}\mathbf E\{m^{d/(2r+d)}h_{D_1}^{-d^2/(2r+d)}\}.
\end{eqnarray*}
Thus
\begin{eqnarray*}
        \mathbf E\{\| \tilde{f}_{\tilde{h}} -f_\rho\}\|^2_\rho\}
        &\leq&
        c_0^2m^{-2r/(2r+d)}\mathbf E\{h_{D_1}^{2rd/(2r+d)}\}
            +
            (c_0^2+2)\mathbf E\{h_{D_1}^{2r}\}\\
            &+&
            8(c_3+2c_4^2)M^2(mn)^{-1}m^{d/(2r+d)}\mathbf E\{h_{D_1}^{-d^2/(2r+d)}\}.
\end{eqnarray*}
To bound $\mathbf E\{h_{D_1}^{2rd/(2r+d)}\}$, we note that for
arbitrary $\varepsilon>0$, there holds
$$
         \mathbf P\{h_{D_1}>\varepsilon\}
         =\mathbf P\{\max_{x\in\mathcal X}\min_{X_i\in D_1}\|x-X_i\|>\varepsilon\}
         \leq  \max_{x\in\mathcal X}
         \mathbf E\{(1-\rho_X(B_\varepsilon(x)))^n\}.
$$
Let $t_1,\dots,t_l$ be the quasi-uniform points  of $\mathcal X$.
Then it follows from \citep[P.93]{gyorfi2006distribution} that
$
        \mathbf P\{h_{D_1}>\varepsilon\}\leq
        \frac{1}{n\varepsilon^d}.
$
Then, we have
\begin{eqnarray*}
    &&
    \mathbf E\{h_{D_1}^{2rd/(2r+d)}\}
    =
    \int_0^\infty\mathbf
    P\{h_{D_1}^{2rd/(2r+d)}>\varepsilon\}d\varepsilon
     =
    \int_0^\infty\mathbf
    P\{h_{D_1}>\varepsilon^{(2r+d)/(2rd)}\}d\varepsilon\\
    &\leq&
    \int_0^{n^{-2r/(2r+d)}}1d\varepsilon+
    \int_{n^{-2r/(2r+d)}}^\infty\mathbf
    P\{h_{D_1}>\varepsilon^{(2r+d)/(2rd)}\}d\varepsilon\\
    &\leq&
    n^{-2r/(2r+d)}+\frac1n\int_{n^{-2r/(2r+d)}}^\infty\varepsilon^{-(2r+d)/(2r)}d\varepsilon
    \leq
    \frac{2r+d}dn^{-2r/(2r+d)}.
\end{eqnarray*}
To bound $\mathbf E\{h_{D_1}^{2r}\}$, we can use the above method
again and $r<d/2$ to derive
$
        \mathbf E\{h_{D_1}^{2r}\}
        \leq 4rd^{-1}n^{-2r/d}.
$
To bound $\mathbf E\{h_{D_1}^{-d^2/(2r+d)}\}$, we use the fact
 $h_{D_1}\geq n^{-1/d}$ almost surely to obtain
$
         \mathbf E\{h_{D_1}^{-d^2/(2r+d)}\}\leq n^{d/(2r+d)}.
$
Hence
$$
        \mathbf E\{\| \tilde{f}_{\tilde{h}} -f_\rho\}\|^2_\rho\}
        \leq
        \left(\frac{c_0^2(2r+d)}d+8(c_3+2c_4^2)M^2\right)N^{-2r/(2r+d)}
            +
            \frac{4r(c_0^2+2)}{d}n^{-2r/d}.
$$
Since
$$
          m\leq
          \left(\frac{ c_0^2(2r+d)+8d(c_3+2c_4^2)M^2}{4r(c_0^2+2)}
          \right)^{d/(2r)}N^{2r/(2r+d)},
$$
we have
$$
        \mathbf E\{\| \tilde{f}_{\tilde{h}} -f_\rho\}\|^2_\rho\}
        \leq
        2\left(\frac{c_0^2(2r+d)}d+8(c_3+2c_4^2)M^2\right)N^{-2r/(2r+d)}
$$
which finishes the proof of (\ref{theorem3}). $\square$

\subsection{Proof of Theorem 4}

{\bf Proof.} From the definition, it follows that
$$
         \hat{f_h}(x)=\sum_{j=1}^m\frac{I_{\{B_h(x)\cap D_j\neq
         \varnothing\}}}{\sum_{j=1}^mI_{\{B_h(x)\cap D_j\neq
         \varnothing\}}}\sum_{(X_i^j,Y_i^j)\in D_j}W_{h,X_{i}^j}(x)Y_{i}^j.
$$
We then use Theorem \ref{THEOREM:SUFFICIENT CONDITION FOR LOCAL
ESTIMATE} to consider a new local estimate with
$$
      W^*_{h,X_{i}^j}(x)=\frac{I_{\{B_h(x)\cap D_j\neq
         \varnothing\}}W_{h,X_{i}^j}(x)}{\sum_{j=1}^mI_{\{B_h(x)\cap D_j\neq
         \varnothing\}}}.
$$
We first prove (A) holds. To this end, we have
\begin{eqnarray*}
         \mathbf E\left\{\sum_{j=1}^m\sum_{(X_i^j,Y_i^j)\in
           D_j}(W^*_{h,X_i^j}(X))^2\right\}
           &\leq&
           \mathbf E\left\{\sum_{j=1}^m\sum_{(X_i^j,Y_i^j)\in
           D_j,X_{i}^j\in B_h(X)}(W^*_{h,X_i^j}(X))^2 \right\}\\
           &+&
           \mathbf E\left\{\sum_{j=1}^m\sum_{(X_i^j,Y_i^j)\in
           D_j,X_{i}^j\notin B_h(X)}(W^*_{h,X_i^j}(X))^2\right\},
\end{eqnarray*}
where we define $\sum_{\varnothing}=0$. To bound the first term in
the right part of the above inequality, it is easy to see that if
$I_{\{X_{i}^j\in B_h(X)\}}=1$, then $I_{\{B_h(X)\cap D_j\neq
         \varnothing\}}=1$, vice versa. Thus, it follows from (C)
         that
\begin{eqnarray*}
           \mathbf E\left\{\sum_{j=1}^m\sum_{(X_i^j,Y_i^j)\in
           D_j,X_{i}^j\in B_h(X)}(W^*_{h,X_i^j}(X))^2  \right\}
           &=&
           \frac1{m^2}\mathbf E\left\{\sum_{j=1}^m\sum_{(X_i^j,Y_i^j)\in
           D_j,X_{i}^j\in B_h(X)}(W_{h,X_i^j}(X))^2  \right\}\\
           &\leq&
           \frac1m\max_{1\leq j\leq m}\mathbf E\left\{\sum_{(X_i^j,Y_i^j)\in
           D_j,X_{i}^j\in B_h(X)}(W_{h,X_i^j}(X))^2  \right\}\\
           &\leq&
           \frac1m\max_{1\leq j\leq m}\mathbf E\left\{\sum_{(X_i^j,Y_i^j)\in
           D_j}(W_{h,X_i^j}(X))^2  \right\}\\
           &\leq&
           \frac{c_{3}}{Nh^d}
\end{eqnarray*}
To bound the second term, we have
\begin{eqnarray*}
           &&\mathbf E\left\{\sum_{j=1}^m\sum_{(X_i^j,Y_i^j)\in
           D_j,X_{i}^j\notin B_h(X)}(W^*_{h,X_i^j}(X))^2\right\}\\
           &&=
           \mathbf E\left\{\sum_{j=1}^m\sum_{(X_i^j,Y_i^j)\in
           D_j,X_{i}^j\notin B_h(X)}\left(\frac{I_{\{B_h(X)\cap D_j\neq
         \varnothing\}}W_{h,X_{i}^j}(X)}{\sum_{j=1}^mI_{\{B_h(X)\cap D_j\neq
         \varnothing\}}}\right)^2\right\}
\end{eqnarray*}
At first, the same method as that in the proof of Theorem
\ref{THEOREM:SUFFICIENT CONDITION FOR LOCAL ESTIMATE} yields that
$
       \mathbf E\{B_h(X)\cap D=\varnothing\}\leq \frac{4}{Nh^d}.
$
Therefore, we have
\begin{eqnarray*}
          &&\mathbf E\left\{\sum_{j=1}^m\sum_{(X_i^j,Y_i^j)\in
           D_j,X_{i}^j\notin B_h(X)}\left(\frac{I_{\{B_h(X)\cap D_j\neq
         \varnothing\}}W_{h,X_{i}^j}(X)}{\sum_{j=1}^mI_{\{B_h(X)\cap D_j\neq
         \varnothing\}}}\right)^2\right\}\\
         &\leq&
         \frac{4}{Nh^d}+m\max_{1\leq j\leq m}\mathbf E\left\{\sum_{(X_i^j,Y_i^j)\in
           D_j}\left(W_{h,X_{i}^j}(X)I_{\|X-X_i^j\|>h}\right)\right\}\\
           &\leq&
           \frac{4+c_{3}+c_5}{Nh^d}.
\end{eqnarray*}
Now, we turn to prove (B) holds. This can be deduced directly by
using the similar method as the last inequality and the condition
(E). That is,
$$
      \mathbf
      E\left\{ \sum_{j=1}^m
      \sum_{(X_i^j,Y_i^j)\in D_j}|W^*_{h,X^j_i}(X)|I_{\{\|X-X_i\|>h\}}\right\}
      \leq
       \frac{c_{5}}{\sqrt{Nh^d}}.
$$
Then Theorem \ref{THEOREM: DLAE} follows from Theorem
\ref{THEOREM:SUFFICIENT CONDITION FOR LOCAL ESTIMATE}. $\square$

\section{Conclusion}\label{section6}
In this paper, we combined  the  divide and conquer strategy with
local average regression to provide a new method called
average-mixture local average regression (AVM-LAR) to  attack the
massive data regression problems. We found that the estimate
obtained by AVM-LAR can achieve the minimax learning rate under a strict
restriction concerning $m$. We then proposed two variants of AVM-LAR to either lessen the restriction or remove it. Theoretical analysis and
simulation studies confirmed our assertions.

We discuss here three interesting  topics for future study.
Firstly, LAR cannot handle the high-dimensional data due to the
curse of
dimensionality~\citep{gyorfi2006distribution,fan2000prospects}. How
to design variants of AVM-LAR to overcome this hurdle can be accommodated as a desirable
research topic. Secondly, we have justified that applying the divide and conquer strategy on the LARs does not degenerate the order of learning rate under mild conditions. However, we did not show there is no loss in the constant factor. Discussing the constant factor of the optimal learning rate is an interesting project. Finally, equipping  other nonparametric methods
(e.g., \citet{fan1994censored,gyorfi2006distribution,tsybakov2008introduction}) with the  divide and conquer strategy can be taken into consideration for massive data analysis. For example, \citet{cheng2015computational} have discussed that how to appropriately apply the divide and conquer strategy to the smoothing spline method.

%
%
%
%
%


\bibliographystyle{asa}
\bibliography{distributedlearning}

\begin{thebibliography}{25}
\newcommand{\enquote}[1]{``#1''}
\expandafter\ifx\csname natexlab\endcsname\relax\def\natexlab#1{#1}\fi

\bibitem[{Battey et~al.(2015)Battey, Fan, Liu, Lu, and
  Zhu}]{battey2015distributed}
Battey, H., Fan, J., Liu, H., Lu, J., and Zhu, Z. (2015), \enquote{Distributed
  Estimation and Inference with Statistical Guarantees,} \textit{arXiv preprint
  arXiv:1509.05457}.

\bibitem[{Biau et~al.(2010)Biau, Cadre, Rouviere, et~al.}]{biau2010statistical}
Biau, G., Cadre, B., Rouviere, L., et~al. (2010), \enquote{Statistical analysis
  of k-nearest neighbor collaborative recommendation,} \textit{The Annals of
  Statistics}, 38, 1568--1592.

\bibitem[{Cheng and Shang(2015)}]{cheng2015computational}
Cheng, G. and Shang, Z. (2015), \enquote{Computational Limits of
  Divide-and-Conquer Method,} \textit{arXiv preprint arXiv:1512.09226}.

\bibitem[{Dwork and Smith(2009)}]{dwork2010differential}
Dwork, C. and Smith, A. (2009), \enquote{Differential privacy for statistics:
  What we know and what we want to learn,} \textit{Journal of Privacy and
  Confidentiality}, 1, 135--154.

\bibitem[{Fan(2000)}]{fan2000prospects}
Fan, J. (2000), \enquote{Prospects of nonparametric modeling,} \textit{Journal
  of the American Statistical Association}, 95, 1296--1300.

\bibitem[{Fan and Gijbels(1994)}]{fan1994censored}
Fan, J. and Gijbels, I. (1994), \enquote{Censored regression: local linear
  approximations and their applications,} \textit{Journal of the American
  Statistical Association}, 89, 560--570.

\bibitem[{Guha et~al.(2012)Guha, Hafen, Rounds, Xia, Li, Xi, and
  Cleveland}]{rhipe2012}
Guha, S., Hafen, R., Rounds, J., Xia, J., Li, J., Xi, B., and Cleveland, W.~S.
  (2012), \enquote{Large complex data: divide and recombine (d\&r) with rhipe,}
  \textit{Stat}, 1, 53--67.

\bibitem[{Gy{\"o}rfi et~al.(2002)Gy{\"o}rfi, Kohler, Krzyzak, and
  Walk}]{gyorfi2006distribution}
Gy{\"o}rfi, L., Kohler, M., Krzyzak, A., and Walk, H. (2002), \textit{A
  distribution-free theory of nonparametric regression}, Springer Science \&
  Business Media.

\bibitem[{Kato(2012)}]{Kato2012}
Kato, K. (2012), \enquote{Weighted Nadaraya--Watson estimation of conditional
  expected shortfall,} \textit{Journal of Financial Econometrics}, 10,
  265--291.

\bibitem[{Kaul et~al.(2013)Kaul, Yang, and Jensen}]{kaul2013building}
Kaul, M., Yang, B., and Jensen, C.~S. (2013), \enquote{Building accurate 3d
  spatial networks to enable next generation intelligent transportation
  systems,} in \textit{Mobile Data Management (MDM), 2013 IEEE 14th
  International Conference on}, IEEE, vol.~1, pp. 137--146.

\bibitem[{Kramer et~al.(2010)Kramer, Satzger, and L{\"a}ssig}]{kramer2010power}
Kramer, O., Satzger, B., and L{\"a}ssig, J. (2010), \enquote{Power prediction
  in smart grids with evolutionary local kernel regression,} in \textit{Hybrid
  Artificial Intelligence Systems}, Springer, pp. 262--269.

\bibitem[{Li et~al.(2013)Li, Lin, and Li}]{li2013statistical}
Li, R., Lin, D.~K., and Li, B. (2013), \enquote{Statistical inference in
  massive data sets,} \textit{Applied Stochastic Models in Business and
  Industry}, 29, 399--409.

\bibitem[{Mcdonald et~al.(2009)Mcdonald, Mohri, Silberman, Walker, and
  Mann}]{Mann2009}
Mcdonald, R., Mohri, M., Silberman, N., Walker, D., and Mann, G.~S. (2009),
  \enquote{Efficient large-scale distributed training of conditional maximum
  entropy models,} in \textit{Advances in Neural Information Processing
  Systems}, pp. 1231--1239.

\bibitem[{Stone(1977)}]{stone1977consistent}
Stone, C.~J. (1977), \enquote{Consistent nonparametric regression,} \textit{The
  annals of statistics}, 595--620.

\bibitem[{Stone(1980)}]{stone1980optimal}
--- (1980), \enquote{Optimal rates of convergence for nonparametric
  estimators,} \textit{The annals of Statistics}, 1348--1360.

\bibitem[{Stone(1982)}]{stone1982optimal}
--- (1982), \enquote{Optimal global rates of convergence for nonparametric
  regression,} \textit{The annals of statistics}, 1040--1053.

\bibitem[{Takeda et~al.(2007)Takeda, Farsiu, and Milanfar}]{takeda2007kernel}
Takeda, H., Farsiu, S., and Milanfar, P. (2007), \enquote{Kernel regression for
  image processing and reconstruction,} \textit{Image Processing, IEEE
  Transactions on}, 16, 349--366.

\bibitem[{Tavares et~al.(2009)Tavares, Zsigraiova, Semiao, and
  Carvalho}]{tavares2009optimisation}
Tavares, G., Zsigraiova, Z., Semiao, V., and Carvalho, M. d.~G. (2009),
  \enquote{Optimisation of MSW collection routes for minimum fuel consumption
  using 3D GIS modelling,} \textit{Waste Management}, 29, 1176--1185.

\bibitem[{Tsybakov(2008)}]{tsybakov2008introduction}
Tsybakov, A.~B. (2008), \textit{Introduction to nonparametric estimation},
  Springer Science \& Business Media.

\bibitem[{Wang et~al.(2015)Wang, Chen, Schifano, Wu, and
  Yan}]{wang2015statistical}
Wang, C., Chen, M.-H., Schifano, E., Wu, J., and Yan, J. (2015),
  \enquote{Statistical Methods and Computing for Big Data,} \textit{arXiv
  preprint arXiv:1502.07989}.

\bibitem[{Wendland(2004)}]{Wendland2005}
Wendland, H. (2004), \textit{Scattered data approximation}, vol.~17, Cambridge
  university press.

\bibitem[{Wu et~al.(2014)Wu, Zhu, Wu, and Ding}]{Wu2014}
Wu, X., Zhu, X., Wu, G.-Q., and Ding, W. (2014), \enquote{Data mining with big
  data,} \textit{Knowledge and Data Engineering, IEEE Transactions on}, 26,
  97--107.

\bibitem[{Zhang et~al.(2015)Zhang, Duchi, and Wainwright}]{Zhang2014}
Zhang, Y., Duchi, J.~C., and Wainwright, M.~J. (2015), \enquote{Divide and
  conquer kernel ridge regression: A distributed algorithm with minimax optimal
  rates,} \textit{Journal of Machine Learning Research, to appear}.

\bibitem[{Zhang et~al.(2013)Zhang, Wainwright, and Duchi}]{Zhang2013}
Zhang, Y., Wainwright, M.~J., and Duchi, J.~C. (2013),
  \enquote{Communication-efficient algorithms for statistical optimization,}
  \textit{Journal of Machine Learning Research}, 14, 3321--3363.

\bibitem[{Zhou et~al.(2014)Zhou, Chawla, Jin, and Williams}]{Zhou2014}
Zhou, Z., Chawla, N., Jin, Y., and Williams, G. (2014), \enquote{Big Data
  Opportunities and Challenges: Discussions from Data Analytics Perspectives
  [Discussion Forum],} \textit{Computational Intelligence Magazine, IEEE}, 9,
  62--74.

\end{thebibliography}
\end{document}